\documentclass[10pt,twocolumn,letterpaper]{article}

\usepackage{iccv}
\usepackage{times}
\usepackage{epsfig}
\usepackage{graphicx}
\usepackage{amsmath}
\usepackage{amssymb}

\usepackage{bm}
\usepackage{subcaption}
\usepackage{algorithm}
\usepackage{algorithmic}
\usepackage{bbm}
\usepackage{authblk}
\usepackage{xcolor}
\usepackage[breaklinks=true,bookmarks=false,colorlinks,letterpaper=true]{hyperref}

\iccvfinalcopy 

\def\iccvPaperID{2490} 

\ificcvfinal\fi
\begin{document}

\title{Depth Completion from Sparse LiDAR Data with Depth-Normal Constraints }

\author{Yan Xu$^{1,2,3}$~~~~Xinge Zhu$^2$~~~~Jianping Shi$^1$~~~~Guofeng Zhang$^3$~~~~Hujun Bao$^3$~~~~Hongsheng Li$^2$ \\
$^1$SenseTime Research~~~~$^2$The Chinese University of Hong Kong\\
$^3$State Key Lab of CAD\&CG, Zhejiang University

}

\begin{figure*}
\vspace*{-2cm}
\noindent\makebox[1\linewidth]{\includegraphics[page=1,trim={0 0 1cm 0},clip, width=1.2\textwidth]{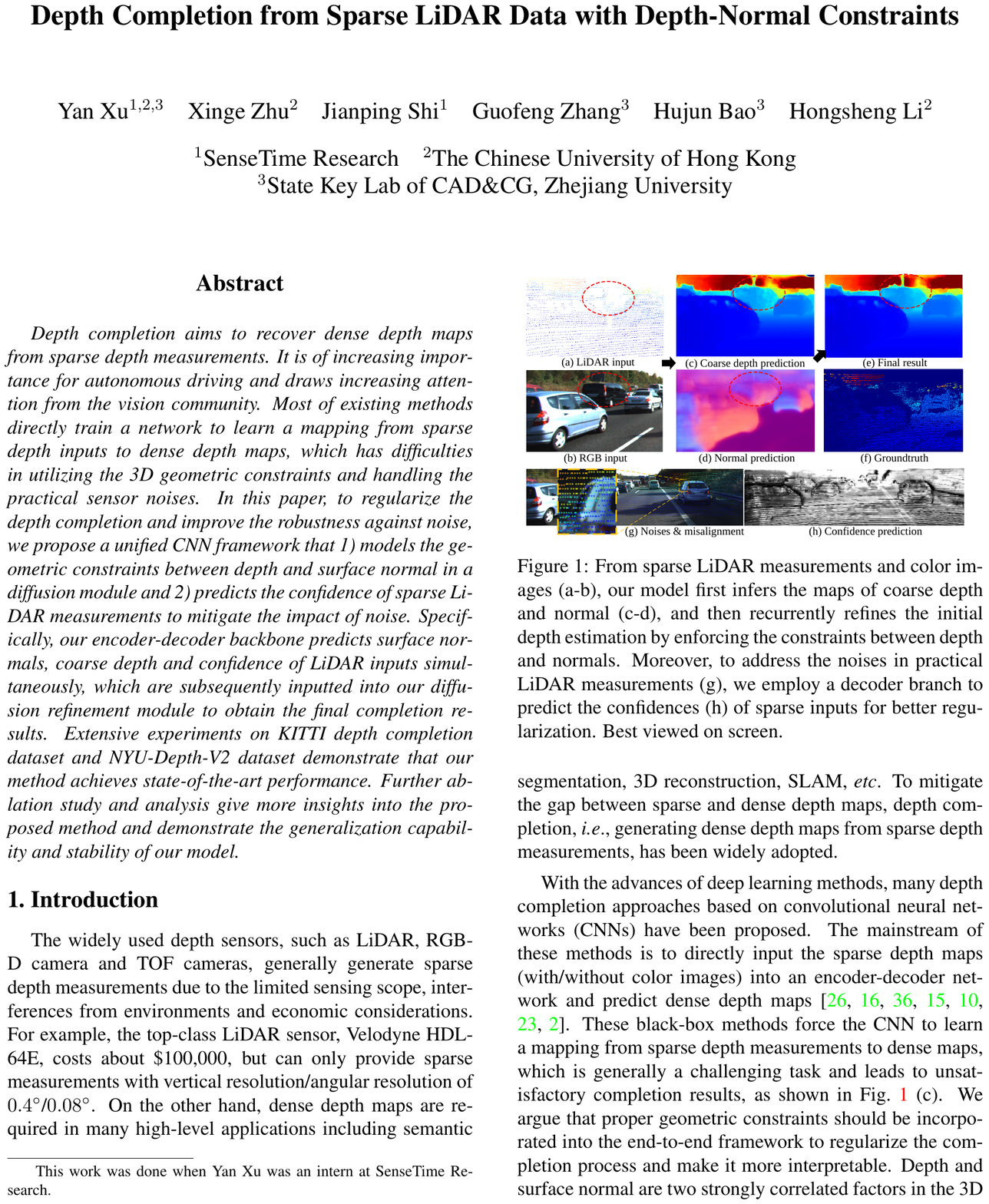}}
\end{figure*}

\begin{figure*}
    \vspace*{-2cm}
    \noindent\makebox[1\linewidth]{\includegraphics[page=2,trim={0 0 1cm 0},clip, width=1.2\textwidth]{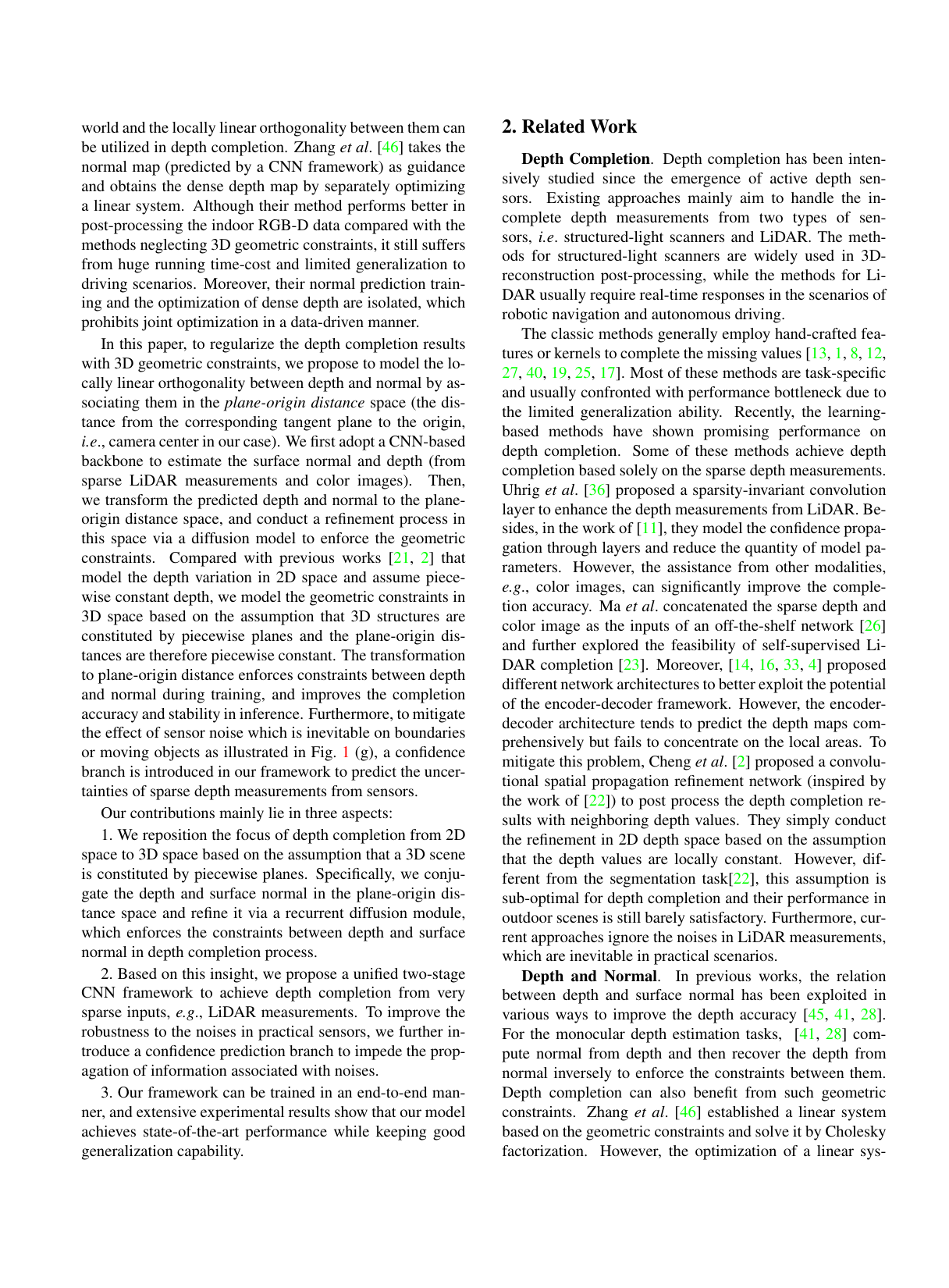}}
    \end{figure*}

\begin{figure*}
    \vspace*{-2cm}
    \noindent\makebox[1\linewidth]{\includegraphics[page=3,trim={0 0 1cm 0},clip, width=1.2\textwidth]{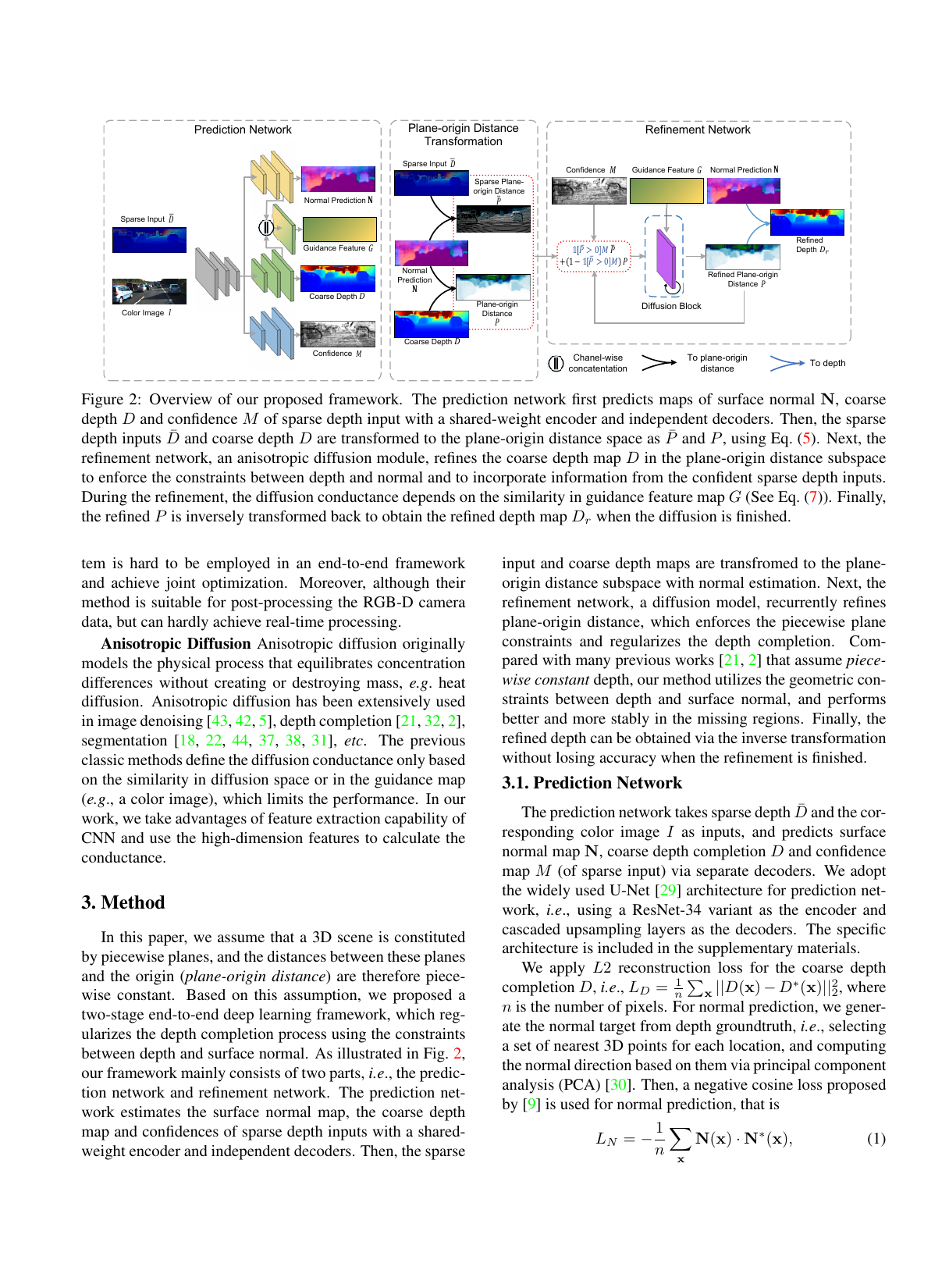}}
\end{figure*}

\begin{figure*}
    \vspace*{-2cm}
    \noindent\makebox[1\linewidth]{\includegraphics[page=4,trim={0 0 1cm 0},clip, width=1.2\textwidth]{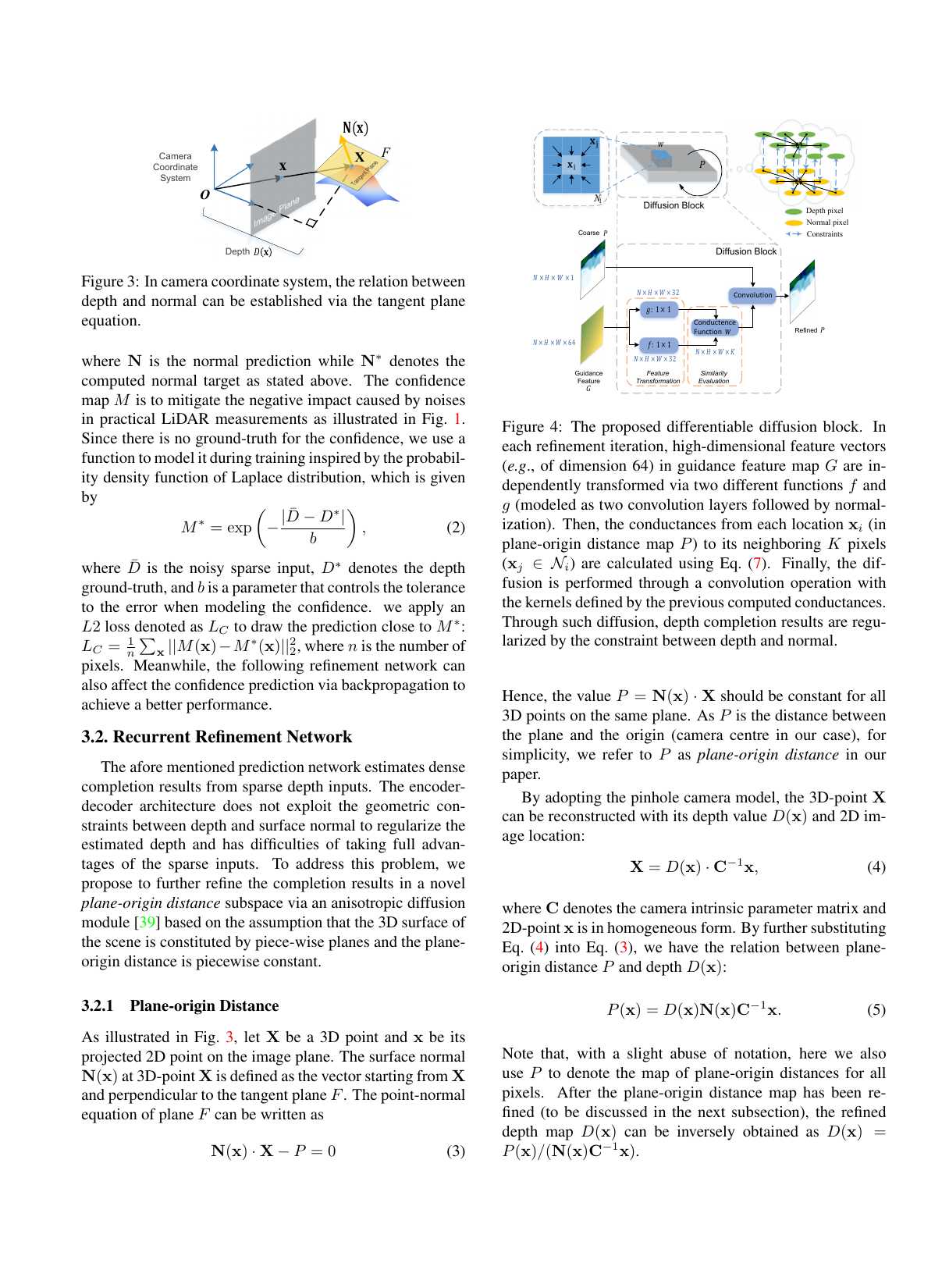}}
\end{figure*}

\begin{figure*}
    \vspace*{-2cm}
    \noindent\makebox[1\linewidth]{\includegraphics[page=5,trim={0 0 1cm 0},clip, width=1.2\textwidth]{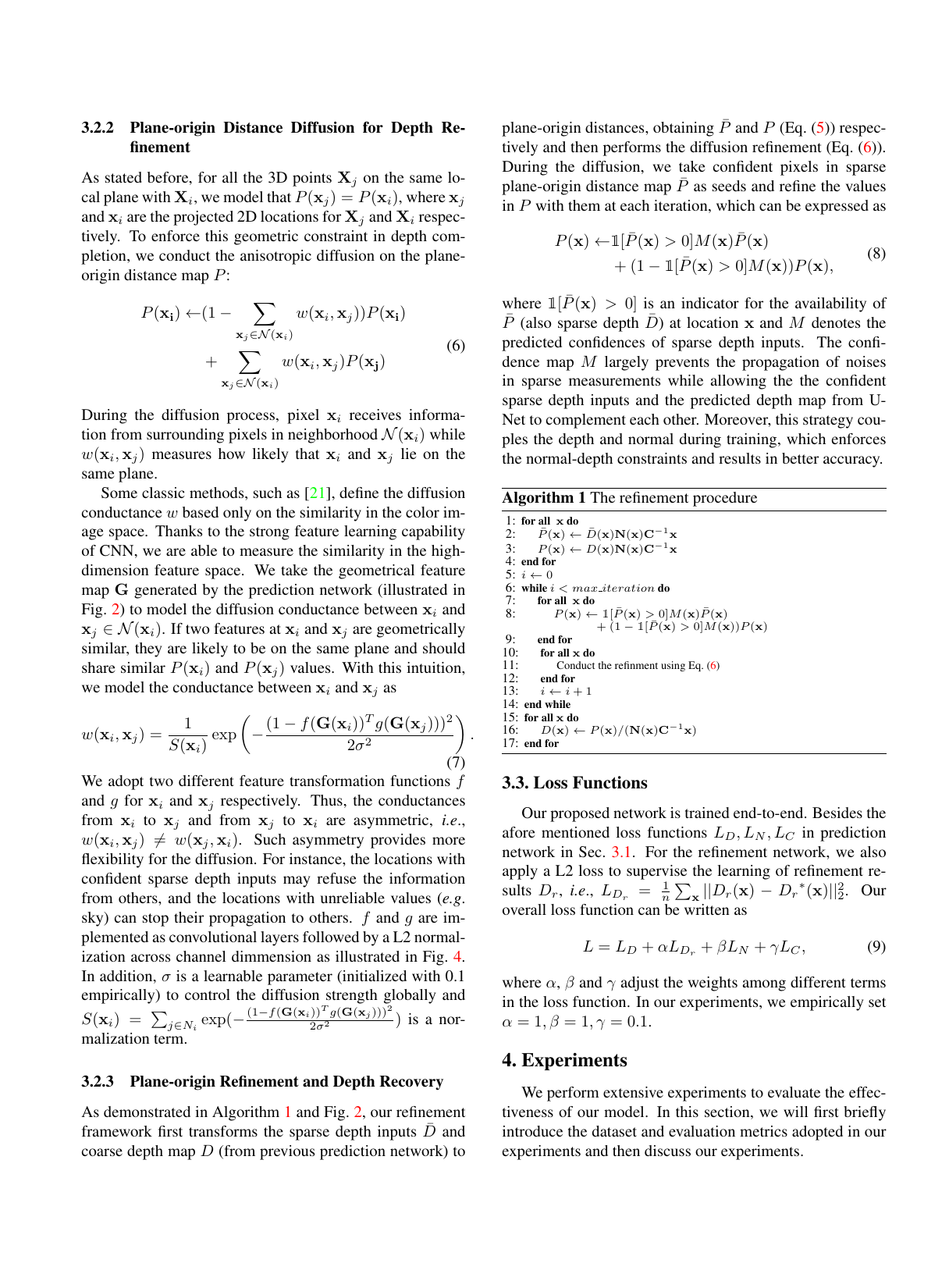}}
\end{figure*}

\begin{figure*}
    \vspace*{-2cm}
    \noindent\makebox[1\linewidth]{\includegraphics[page=6,trim={0 0 1cm 0},clip, width=1.2\textwidth]{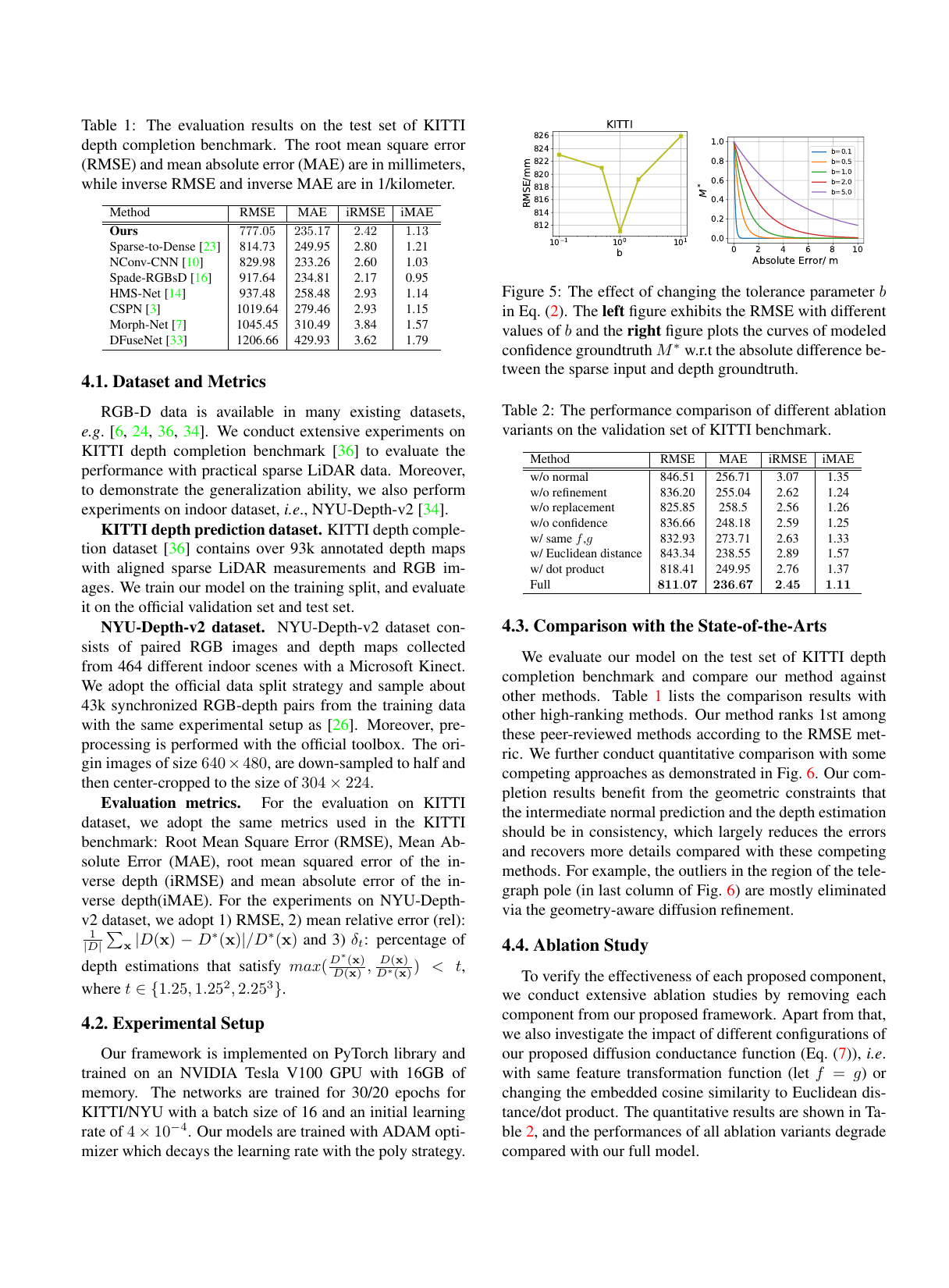}}
\end{figure*}

\begin{figure*}
    \vspace*{-2cm}
    \noindent\makebox[1\linewidth]{\includegraphics[page=7,trim={0 0 1cm 0},clip, width=1.2\textwidth]{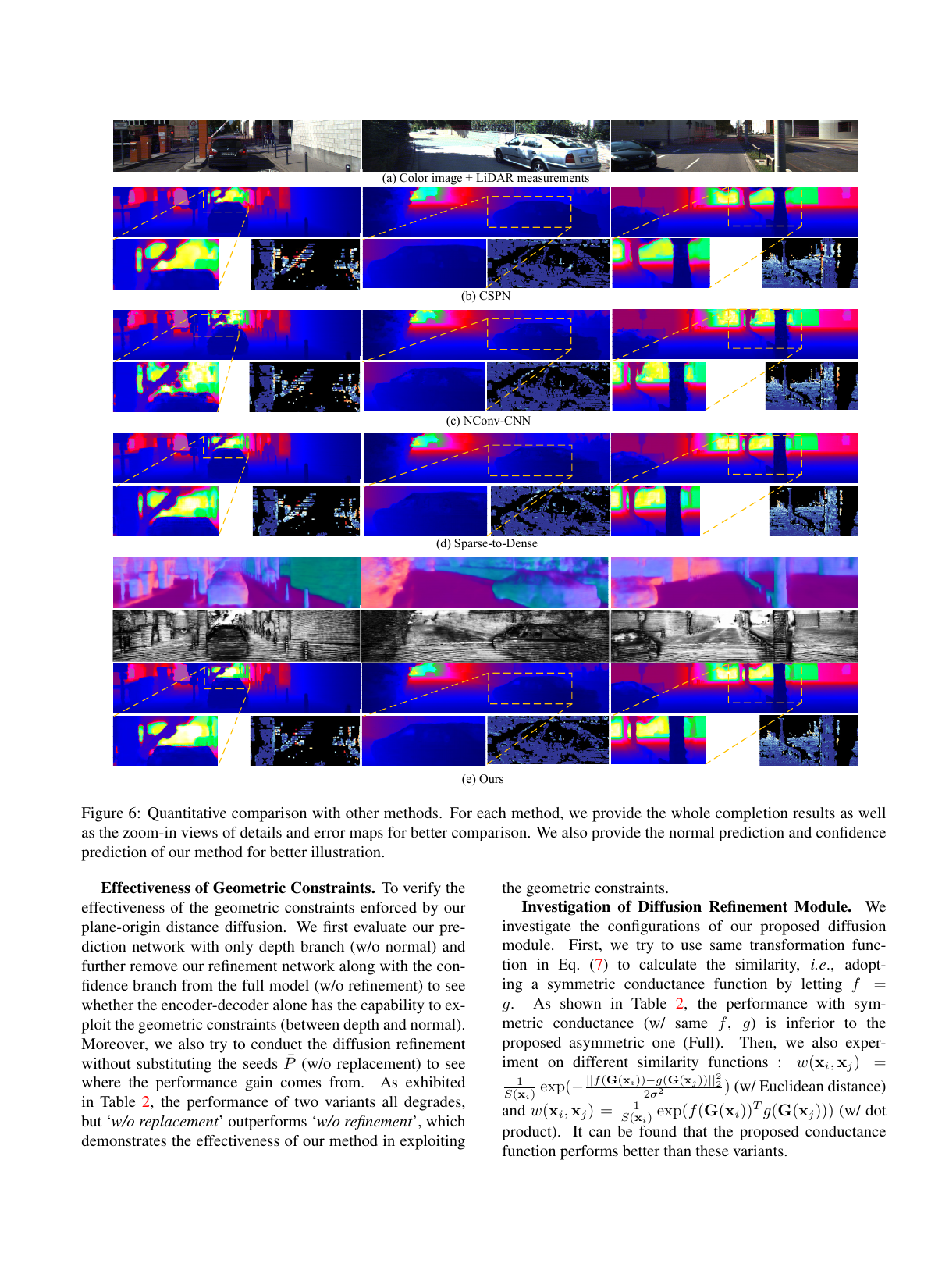}}
\end{figure*}

\begin{figure*}
    \vspace*{-2cm}
    \noindent\makebox[1\linewidth]{\includegraphics[page=8,trim={0 0 1cm 0},clip, width=1.2\textwidth]{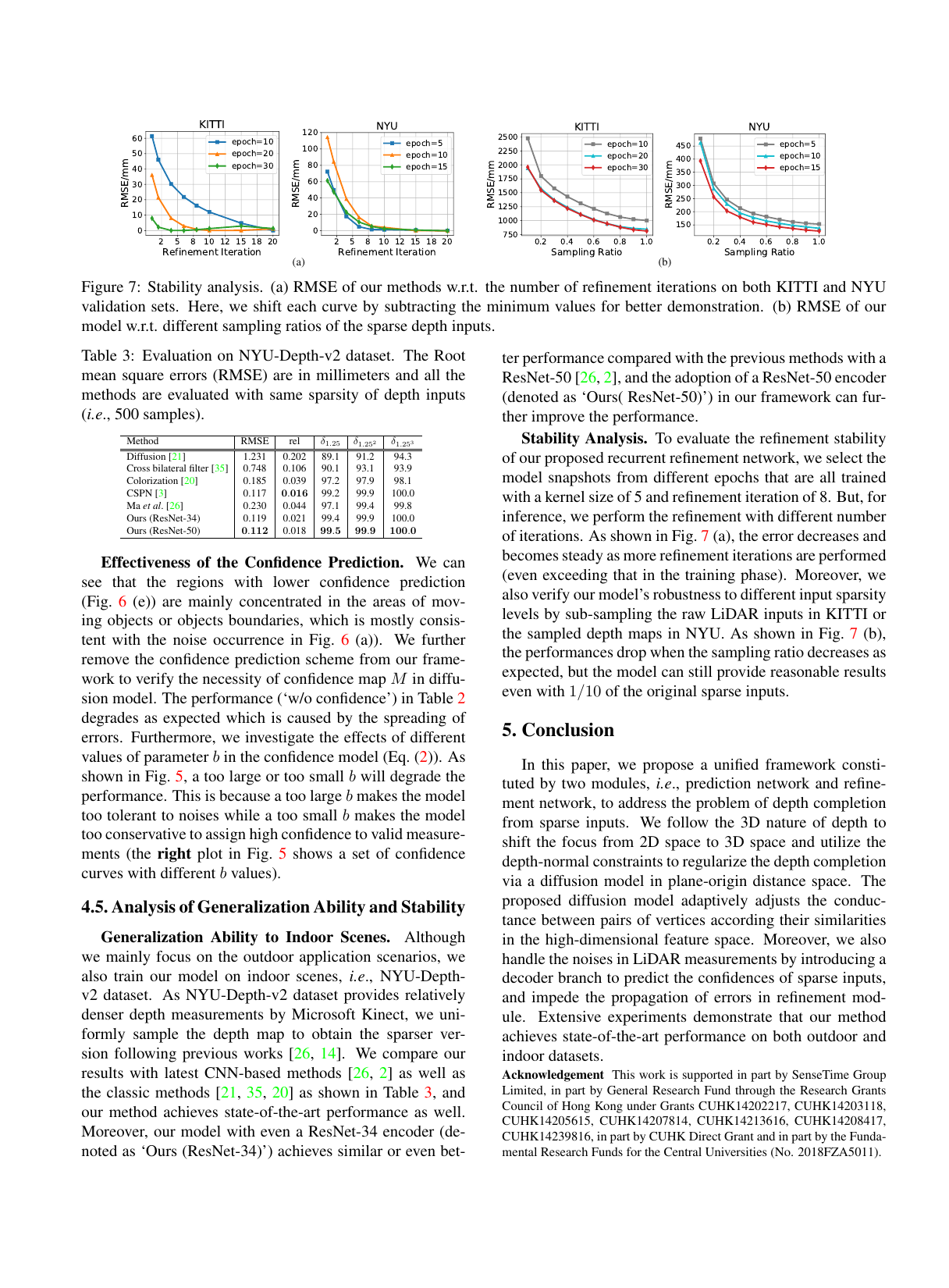}}
\end{figure*}

\begin{figure*}
    \vspace*{-2cm}
    \noindent\makebox[1\linewidth]{\includegraphics[page=9,trim={0 0 1cm 0},clip, width=1.2\textwidth]{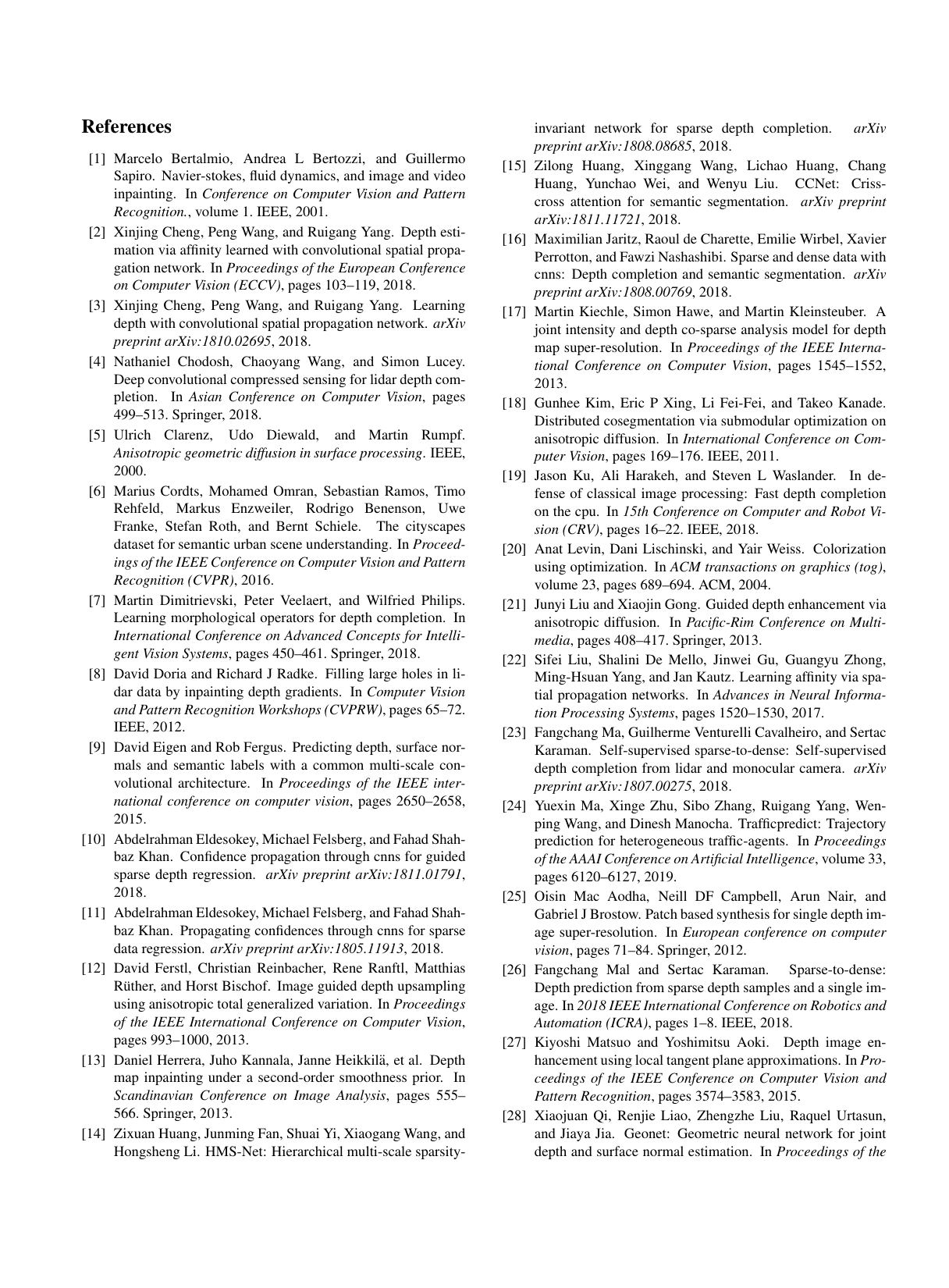}}
\end{figure*}

\begin{figure*}
    \vspace*{-2cm}
    \noindent\makebox[1\linewidth]{\includegraphics[page=10,trim={0 0 1cm 0},clip, width=1.2\textwidth]{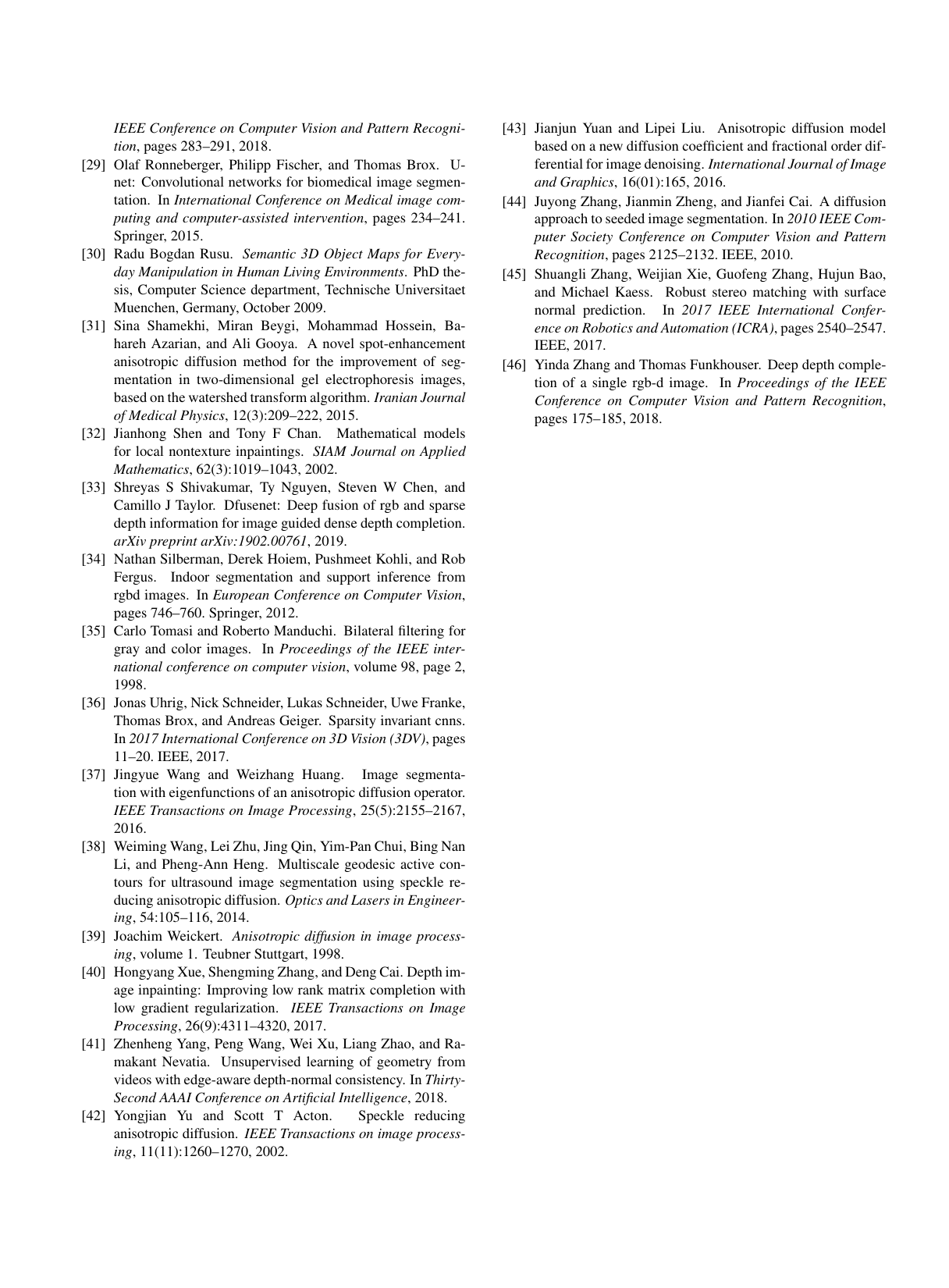}}
\end{figure*}

\clearpage
{
\fontsize{1}{1}{\selectfont
\textcolor{white}{
\begin{abstract}
Sorry about this little trick and hope it would work, since I do not have much time to neatten my original source code. 
depth completion aims to recover dense depth maps
from sparse depth measurements. It is of increasing importance for autonomous driving and draws increasing attention from the vision community. Most of existing methods
directly train a network to learn a mapping from sparse
depth inputs to dense depth maps, which has difficulties
in utilizing the 3D geometric constraints and handling the
practical sensor noises. In this paper, to regularize the
depth completion and improve the robustness against noise,
we propose a unified CNN framework that 1) models the geometric constraints between depth and surface normal in a
\end{abstract}
\section{Introduction}
Depth completion aims to recover dense depth maps
from sparse depth measurements. It is of increasing importance for autonomous driving and draws increasing attention from the vision community. Most of existing methods
directly train a network to learn a mapping from sparse
depth inputs to dense depth maps, which has difficulties
in utilizing the 3D geometric constraints and handling the
practical sensor noises. In this paper, to regularize the
depth completion and improve the robustness against noise,
we propose a unified CNN framework that 1) models the geometric constraints between depth and surface normal in a
diffusion module and 2) predicts the confidence of sparse LiDAR measurements to mitigate the impact of noise. Specifically, our encoder-decoder backbone predicts surface normals, coarse depth and confidence of LiDAR inputs simultaneously, which are subsequently inputted into our diffusion refinement module to obtain the final completion results. Extensive experiments on KITTI depth completion
dataset and NYU-Depth-V2 dataset demonstrate that our
method achieves state-of-the-art performance. Further ablation study and analysis give more insights into the proposed method and demonstrate the generalization capability and stability of our model.
Depth Completion. Depth completion has been intensively studied since the emergence of active depth sensors. Existing approaches mainly aim to handle the incomplete depth measurements from two types of sensors, i.e. structured-light scanners and LiDAR. The methods for structured-light scanners are widely used in 3Dreconstruction post-processing, while the methods for LiDAR usually require real-time responses in the scenarios of
robotic navigation and autonomous driving.
The classic methods generally employ hand-crafted features or kernels to complete the missing values [13, 1, 8, 12,
27, 40, 19, 25, 17]. Most of these methods are task-specific
and usually confronted with performance bottleneck due to
the limited generalization ability. Recently, the learningbased methods have shown promising performance on
depth completion. Some of these methods achieve depth
completion based solely on the sparse depth measurements.
Uhrig et al. [36] proposed a sparsity-invariant convolution
layer to enhance the depth measurements from LiDAR. Besides, in the work of [11], they model the confidence propagation through layers and reduce the quantity of model parameters. However, the assistance from other modalities,
e.g., color images, can significantly improve the completion accuracy. Ma et al. concatenated the sparse depth and
color image as the inputs of an off-the-shelf network [26]
and further explored the feasibility of self-supervised LiDAR completion [23]. Moreover, [14, 16, 33, 4] proposed
different network architectures to better exploit the potential
of the encoder-decoder framework. However, the encoderdecoder architecture tends to predict the depth maps comprehensively but fails to concentrate on the local areas. To
mitigate this problem, Cheng et al. [2] proposed a convolutional spatial propagation refinement network (inspired by
the work of [22]) to post process the depth completion results with neighboring depth values. They simply conduct
the refinement in 2D depth space based on the assumption
that the depth values are locally constant. However, different from the segmentation task[22], this assumption is
sub-optimal for depth completion and their performance in
outdoor scenes is still barely satisfactory. Furthermore, current approaches ignore the noises in LiDAR measurements,
which are inevitable in practical scenarios.
Depth and Normal. In previous works, the relation
between depth and surface normal has been exploited in
various ways to improve the depth accuracy [45, 41, 28].
For the monocular depth estimation tasks, [41, 28] compute normal from depth and then recover the depth from
normal inversely to enforce the constraints between them.
Depth completion can also benefit from such geometric
constraints. Zhang et al. [46] established a linear system
based on the geometric constraints and solve it by Cholesky
factorization. However, the optimization of a linear sys-tem is hard to be employed in an end-to-end framework
and achieve joint optimization. Moreover, although their
method is suitable for post-processing the RGB-D camera
data, but can hardly achieve real-time processing.
Anisotropic Diffusion Anisotropic diffusion originally
models the physical process that equilibrates concentration
differences without creating or destroying mass, e.g. heat
diffusion. Anisotropic diffusion has been extensively used
in image denoising [43, 42, 5], depth completion [21, 32, 2],
segmentation [18, 22, 44, 37, 38, 31], etc. The previous
classic methods define the diffusion conductance only based
on the similarity in diffusion space or in the guidance map
(e.g., a color image), which limits the performance. In our
work, we take advantages of feature extraction capability of
CNN and use the high-dimension features to calculate the
conductanceIn this paper, we assume that a 3D scene is constituted
by piecewise planes, and the distances between these planes
and the origin (plane-origin distance) are therefore piecewise constant. Based on this assumption, we proposed a
two-stage end-to-end deep learning framework, which regularizes the depth completion process using the constraints
between depth and surface normal. As illustrated in Fig. 2,
our framework mainly consists of two parts, i.e., the prediction network and refinement network. The prediction network estimates the surface normal map, the coarse depth
map and confidences of sparse depth inputs with a sharedweight encoder and independent decoders. Then, the sparseThe afore mentioned prediction network estimates dense
completion results from sparse depth inputs. The encoderdecoder architecture does not exploit the geometric constraints between depth and surface normal to regularize the
estimated depth and has difficulties of taking full advantages of the sparse inputs. To address this problem, we
propose to further refine the completion results in a novel
plane-origin distance subspace via an anisotropic diffusion
module [39] based on the assumption that the 3D surface of
the scene is constituted by piece-wise planes and the planeorigin distance is piecewise constant.As stated before, for all the 3D points Xj on the same local plane with Xi
, we model that P(xj ) = P(xi), where xj
and xi are the projected 2D locations for Xj and Xi respectively. To enforce this geometric constraint in depth completion, we conduct the anisotropic diffusion on the planeorigin distance map P:re ✶[P¯(x) > 0] is an indicator for the availability of
P¯ (also sparse depth D¯) at location x and M denotes the
predicted confidences of sparse depth inputs. The confi-
dence map M largely prevents the propagation of noises
in sparse measurements while allowing the the confident
sparse depth inputs and the predicted depth map from UNet to complement each other. Moreover, this strategy couples the depth and normal during training, which enforces
the normal-depth constraints and results in better accuracy.3.2.3 Plane-origin Refinement and Depth Recovery
As demonstrated in Algorithm 1 and Fig. 2, our refinement
framework first transforms the sparse depth inputs D¯ and
coarse depth map D (from previous prediction network) to
plane-origin distances, obtaining P¯ and P (Eq. (5)) respectively and then performs the diffusion refinement (Eq. (6)).
During the diffusion, we take confident pixels in sparse
plane-origin distance map P¯ as seeds and refine the values
in P with them at each iteration, which can be expressed as
P(x) ←✶[P¯(x) > 0]M(x)P¯(x)
+ (1 − ✶[P¯(x) > 0]M(x))P(x),
(8)
where ✶[P¯(x) > 0] is an indicator for the availability of
P¯ (also sparse depth D¯) at location x and M denotes the
predicted confidences of sparse depth inputs. The confi-
dence map M largely prevents the propagation of noises
in sparse measurements while allowing the the confident
sparse depth inputs and the predicted depth map from UNet to complement each other. Moreover, this strategy couples the depth and normal during training, which enforces
the normal-depth constraints and results in better accuracy.
Algorithm 1 The refinement procedure
1: for all x do
2: P¯(x) ← D¯(x)N(x)C−1x
3: P (x) ← D(x)N(x)C−1x
4: end for
5: i ← 0
6: while i < max iteration do
7: for all x do
8: P (x) ← ✶[P¯(x) > 0]M(x)P¯(x)
+ (1 − ✶[P¯(x) > 0]M(x))P (x)
9: end for
10: for all x do
11: Conduct the refinment using Eq. (6)
12: end for
13: i ← i + 1
14: end while
15: for all x do
16: D(x) ← P (x)/(N(x)C−1x)
17: end for
3.3. Loss Functions
Our proposed network is trained end-to-end. Besides the
afore mentioned loss functions LD, LN , LC in prediction
network in Sec. 3.1. For the refinement network, we also
apply a L2 loss to supervise the learning of refinement results Dr, i.e., LDr =
1
n
P
x
||Dr(x) − Dr
∗
(x)||2
2
. Our
overall loss function can be written as
L = LD + αLDr + βLN + γLC , (9)
where α, β and γ adjust the weights among different terms
in the loss function. In our experiments, we empirically set
α = 1, β = 1, γ = 0.1.
4. Experiments
We perform extensive experiments to evaluate the effectiveness of our model. In this section, we will first briefly
introduce the dataset and evaluation metrics adopted in our
experiments and then discuss our experiments4.1. Dataset and Metrics
RGB-D data is available in many existing datasets,
e.g. [6, 24, 36, 34]. We conduct extensive experiments on
KITTI depth completion benchmark [36] to evaluate the
performance with practical sparse LiDAR data. Moreover,
to demonstrate the generalization ability, we also perform
experiments on indoor dataset, i.e., NYU-Depth-v2 [34].
KITTI depth prediction dataset. KITTI depth completion dataset [36] contains over 93k annotated depth maps
with aligned sparse LiDAR measurements and RGB images. We train our model on the training split, and evaluate
it on the official validation set and test set.
NYU-Depth-v2 dataset. NYU-Depth-v2 dataset consists of paired RGB images and depth maps collected
from 464 different indoor scenes with a Microsoft Kinect.
We adopt the official data split strategy and sample about
43k synchronized RGB-depth pairs from the training data
with the same experimental setup as [26]. Moreover, preprocessing is performed with the official toolbox. The origin images of size 640×480, are down-sampled to half and
then center-cropped to the size of 304 × 224.
Evaluation metrics. For the evaluation on KITTI
dataset, we adopt the same metrics used in the KITTI
benchmark: Root Mean Square Error (RMSE), Mean Absolute Error (MAE), root mean squared error of the inverse depth (iRMSE) and mean absolute error of the inverse depth(iMAE). For the experiments on NYU-Depthv2 dataset, we adopt 1) RMSE, 2) mean relative error (rel):
1
|D|
P
x
|D(x) − D∗
(x)|/D∗
(x) and 3) δt: percentage of
depth estimations that satisfy max(
D∗
(x)
D(x)
,
D(x)
D∗(x)
) < t,
where t ∈ {1.25, 1.252
, 2.253}.
4.2. Experimental Setup
Our framework is implemented on PyTorch library and
trained on an NVIDIA Tesla V100 GPU with 16GB of
memory. The networks are trained for 30/20 epochs for
KITTI/NYU with a batch size of 16 and an initial learning
rate of 4 × 10−4
. Our models are trained with ADAM op4. Ablation Study
To verify the effectiveness of each proposed component,
we conduct extensive ablation studies by removing each
component from our proposed framework. Apart from that,
we also investigate the impact of different configurations of
our proposed diffusion conductance function (Eq. (7)), i.e.
with same feature transformation function (let f = g) or
changing the embedded cosine similarity to Euclidean distance/dot product. The quantitative results are shown in Table 2, and the performances of all ablation variants degrade
compared with our full modelEffectiveness of Geometric Constraints. To verify the
effectiveness of the geometric constraints enforced by our
plane-origin distance diffusion. We first evaluate our prediction network with only depth branch (w/o normal) and
further remove our refinement network along with the con-
fidence branch from the full model (w/o refinement) to see
whether the encoder-decoder alone has the capability to exploit the geometric constraints (between depth and normal).
Moreover, we also try to conduct the diffusion refinement
without substituting the seeds P¯ (w/o replacement) to see
where the performance gain comes from. As exhibited
in Table 2, the performance of two variants all degrades,
but ‘w/o replacement’ outperforms ‘w/o refinement’, which
demonstrates the effectiveness of our method in exploiting
the geometric constraints.
Investigation of Diffusion Refinement Module. We
investigate the configurations of our proposed diffusion
module. First, we try to use same transformation function in Eq. (7) to calculate the similarity, i.e., adopting a symmetric conductance function by letting f =
g. As shown in Table 2, the performance with symmetric conductance (w/ same f, g) is inferior to the
proposed asymmetric one (Full). Then, we also experiment on different similarity functions : w(xi
, xj ) =
1
S(xi)
exp(−
||f(G(xi))−g(G(xj ))||2
2
2σ2 ) (w/ Euclidean distance)
and w(xi
, xj ) = 1
S(xi)
exp(f(G(xi))T
g(G(xj ))) (w/ dot
product). It can be found that the proposed conductance
function performs better than these variantsTable 3: Evaluation on NYU-Depth-v2 dataset. The Root
mean square errors (RMSE) are in millimeters and all the
methods are evaluated with same sparsity of depth inputs
(i.e., 500 samples).
Method RMSE rel δ1.25 δ1.252 δ1.253
Diffusion [21] 1.231 0.202 89.1 91.2 94.3
Cross bilateral filter [35] 0.748 0.106 90.1 93.1 93.9
Colorization [20] 0.185 0.039 97.2 97.9 98.1
CSPN [3] 0.117 0.016 99.2 99.9 100.0
Ma et al. [26] 0.230 0.044 97.1 99.4 99.8
Ours (ResNet-34) 0.119 0.021 99.4 99.9 100.0
Ours (ResNet-50) 0.112 0.018 99.5 99.9 100.0
Effectiveness of the Confidence Prediction. We can
see that the regions with lower confidence prediction
(Fig. 6 (e)) are mainly concentrated in the areas of moving objects or objects boundaries, which is mostly consistent with the noise occurrence in Fig. 6 (a)). We further
remove the confidence prediction scheme from our framework to verify the necessity of confidence map M in diffusion model. The performance (‘w/o confidence’) in Table 2
degrades as expected which is caused by the spreading of
errors. Furthermore, we investigate the effects of different
values of parameter b in the confidence model (Eq. (2)). As
shown in Fig. 5, a too large or too small b will degrade the
performance. This is because a too large b makes the model
too tolerant to noises while a too small b makes the model
too conservative to assign high confidence to valid measurements (the right plot in Fig. 5 shows a set of confidence
curves with different b values).
4.5. Analysis of Generalization Ability and Stability
Generalization Ability to Indoor Scenes. Although
we mainly focus on the outdoor application scenarios, we
also train our model on indoor scenes, i.e., NYU-Depthv2 dataset. As NYU-Depth-v2 dataset provides relatively
denser depth measurements by Microsoft Kinect, we uniformly sample the depth map to obtain the sparser version following previous works [26, 14]. We compare our
results with latest CNN-based methods [26, 2] as well as
the classic methods [21, 35, 20] as shown in Table 3, and
our method achieves state-of-the-art performance as well.
Moreover, our model with even a ResNet-34 encoder (denoted as ‘Ours (ResNet-34)’) achieves similar or even better performance compared with the previous methods with a
ResNet-50 [26, 2], and the adoption of a ResNet-50 encoder
(denoted as ‘Ours( ResNet-50)’) in our framework can further improve the performance.
Stability Analysis. To evaluate the refinement stability
of our proposed recurrent refinement network, we select the
model snapshots from different epochs that are all trained
with a kernel size of 5 and refinement iteration of 8. But, for
inference, we perform the refinement with different number
of iterations. As shown in Fig. 7 (a), the error decreases and
becomes steady as more refinement iterations are performed
(even exceeding that in the training phase). Moreover, we
also verify our model’s robustness to different input sparsity
levels by sub-sampling the raw LiDAR inputs in KITTI or
the sampled depth maps in NYU. As shown in Fig. 7 (b),
the performances drop when the sampling ratio decreases as
expected, but the model can still provide reasonable results
even with 1/10 of the original sparse inputs.
5. Conclusion
In this paper, we propose a unified framework constituted by two modules, i.e., prediction network and refinement network, to address the problem of depth completion
from sparse inputs. We follow the 3D nature of depth to
shift the focus from 2D space to 3D space and utilize the
depth-normal constraints to regularize the depth completion
via a diffusion model in plane-origin distance space. The
proposed diffusion model adaptively adjusts the conductance between pairs of vertices according their similarities
in the high-dimensional feature space. Moreover, we also
handle the noises in LiDAR measurements by introducing a
decoder branch to predict the confidences of sparse inputs,
and impede the propagation of errors in refinement module. Extensive experiments demonstrate that our method
achieves state-of-the-art performance on both outdoor and
indoor datasets
\section{Method}
\cite{Cordts2016Cityscapes,RusuDoctoralDissertation,ahn2012depth,cheng2018learning,bertalmio2001navier,chodosh2018deep,doria2012filling,gong2013guided,ku2018defense,muddala2014depth,cheng2018learning,tomasi1998bilateral,clarenz2000anisotropic,herrera2013depth}
}
\textcolor{white}
{
\fontsize{1}{1}{\selectfont
\bibliographystyle{ieee_fullname}
\bibliography{egbib}
}
}
    }
\end{document}